%% file: main.tex
\documentclass{article}

\usepackage{amsmath,amsfonts,amssymb,times,graphicx,natbib,algorithm,algorithmic,hyperref}

\usepackage{booktabs} 

\usepackage[accepted]{whi2020}

\DeclareMathOperator{\relu}{ReLU}

\icmltitlerunning{Rethinking Positive Aggregation and Propagation}

\begin{document}

\twocolumn[
\icmltitle{Rethinking Positive Aggregation and Propagation of Gradients in Gradient-based Saliency Methods}


\icmlsetsymbol{equal}{*}

\begin{icmlauthorlist}
\icmlauthor{Ashkan Khakzar}{tum}
\icmlauthor{Soroosh Baselizadeh}{tum}
\icmlauthor{Nassir Navab}{tum,jhu}
\end{icmlauthorlist}

\icmlaffiliation{tum}{Technical University of Munich, Germany}
\icmlaffiliation{jhu}{Johns Hopkins University, USA}

\icmlcorrespondingauthor{Ashkan Khakzar}{ashkan.khakzar@tum.de}

\icmlkeywords{interpretability, transparency}

\vskip 0.3in
]


\printAffiliationsAndNotice{}

\begin{abstract}
\input{content/abstract}
\end{abstract}

\section{Introduction}
\input{content/intro}

\section{Experimental Setup}\label{sec:exp}
\input{content/experimental_setup}

\section{Positive Aggregation}
\input{content/positive_sum}

\section{Positive Propagation}

\input{content/positive_prop}

\section{Conclusion}
\input{content/conclusion}

\bibliography{references}
\bibliographystyle{icml2020}

\end{document}

%% file: content/abstract.tex
Saliency methods interpret the prediction of a neural network by showing the importance of input elements for that prediction. A popular family of saliency methods utilize gradient information. In this work, we empirically show that two approaches for handling the gradient information, namely positive aggregation, and positive propagation, break these methods. Though these methods reflect visually salient information in the input, they do not explain the model prediction anymore as the generated saliency maps are insensitive to the predicted output and are insensitive to model parameter randomization. Specifically for methods that aggregate the gradients of a chosen layer such as GradCAM++ and FullGrad, exclusively aggregating positive gradients is detrimental. We further support this by proposing several variants of aggregation methods with positive handling of gradient information. For methods that backpropagate gradient information such as LRP, RectGrad, and Guided Backpropagation, we show the destructive effect of exclusively propagating positive gradient information.


%% file: content/intro.tex
\begin{figure*}[t]
    \centering
    \includegraphics[width=0.8\textwidth]{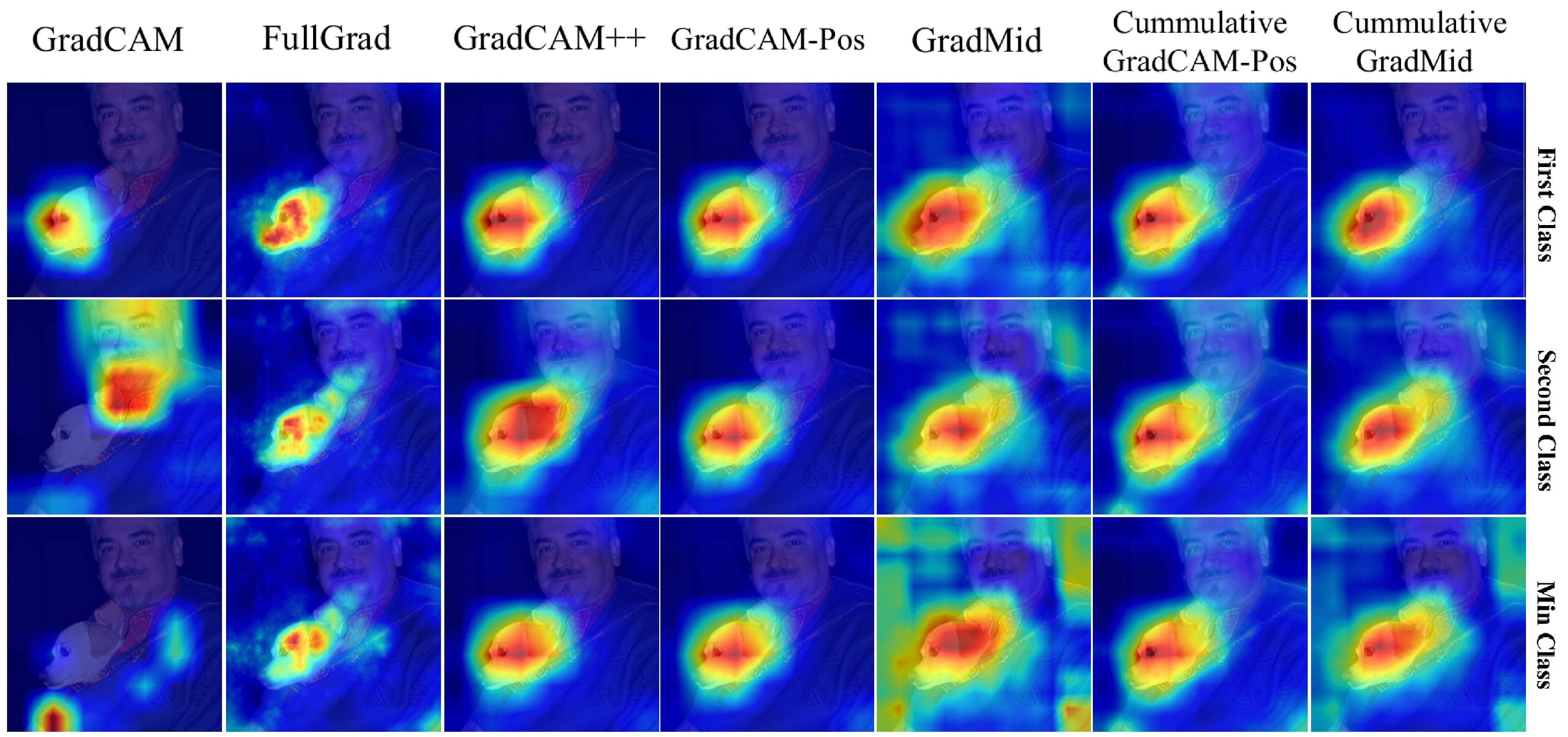}
    \caption{Saliency maps derived from different aggregation methods (columns) for different output class predictions (rows). All methods except GradCAM use positive aggregation and are not class-sensitive.}
    \label{fig:aggregation_all_pointing}
\end{figure*}

The quest for understanding the basis of neural networks’ predictions is gaining momentum as the need for interpretability intensifies especially in sensitive application domains such as medicine, finance, policy, and law. One approach for understanding the predictions is using saliency methods (aka attribution methods) which assign each input element (\textit{e.g.}~pixels) its importance for the corresponding output prediction. 
There exist a plethora of proposed saliency methods, and thus raising the question “which explanation method to trust?”. 
Visual evaluation of saliency maps has been shown to be unreliable as the method can generate human interpretable saliency maps but not explaining model behavior~\cite{Adebayo2018,Kindermans2019,Nie2018,hooker2019benchmark}. One way to evaluate saliency methods is by checking whether they satisfy certain desirable properties. 
Some of these properties are formulated as axioms~\cite{Sundararajan2017,Lundberg2017}, such as efficiency (completeness), dummy (null-player), symmetry, sensitivity, and implementation invariance. It is therefore plausible to theoretically show whether methods abide by the axioms, though there is the caveat that while a method’s formulation satisfies axioms, the required practical assumptions and approximations can break them \cite{sundararajan2019many}. 

It is also possible to formulate desirable properties as experiments (sanity checks) and empirically test whether a method violates those properties. Two important properties that a saliency method is required to satisfy are the sensitivity of the saliency map to the predicted output class~\cite{Nie2018}, aka class-sensitivity, and the sensitivity of the saliency map to model parameter randomization~\cite{Adebayo2018}, i.e. the saliency map for the model before randomizing its weights should be different from the saliency map after the model parameters are randomized. 
Not satisfying one of these two properties signifies that the method cannot explain model behavior. 

In this work, we empirically show that two common approaches for handling gradient information in gradient-based saliency methods, methods that utilize gradient information to explain the model, can result in violating class-sensitivity and randomization-sensitivity. 
We show that in methods which aggregate gradients in a layer, if this aggregation is done only on positive gradients (e.g. via using a ReLU function) the desired properties are not satisfied. We show this phenomenon for FullGrad~\cite{Srinivas2019} and GradCAM++~\cite{Chattopadhay2018}, and other positive aggregation methods that we propose: positive aggregated GradCAM~\cite{Selvaraju2020} for various layers, positive aggregated Gradients for various layers, and a version of both where results from all layers are aggregated.
Another approach that we investigate is inherent in methods that propagate gradient information such as LRP~\cite{Montavon2017,Bach2015} (and Excitation Backpropagation~\cite{zhang2018top} as it is equivalent to LRP-$\alpha1\beta0$), RectGrad~\cite{Kim2020}, and Guided Backpropagation~\cite{Springenberg2015}. We show that, when only positive information is backpropagated, the properties are not satisfied. We investigate these hypotheses empirically via ablative experiments. For sensitivity to model parameter randomization, we use the sanity checks of~\cite{Adebayo2018}. For evaluating class sensitivity we use the class sensitivity metric proposed by~\cite{Rebuffi2020} and the pointing game of\cite{zhang2018top,fong2019understanding}. We also propose the restricted pointing game for evaluating class sensitivity.

\cite{Rebuffi2020} propose a general formulation for aggregation methods, however, do not discuss how the choice of filtering functions (such as $|\cdot|$ or $\relu$) affect them. It is generally observed that these aggregations are not class sensitive.
In this work, we reveal the culprit and also show positive filtering affects the propagation methods.


%% file: content/experimental_setup.tex
The goal in experiments is to evaluate class-sensitivity and sensitivity to model parameter randomization. For this purpose we use the following experiments (we use the VGG-16 network~\cite{simonyan2014very}):
\\
\textbf{Pointing game.} ~\cite{zhang2018top,fong2019understanding}. Saliency maps are generated for each class present in the image. 
For each class, if the maximum value in the generated saliency map is inside the ground truth mask the method has correctly pointed to the class. The accuracy is the ratio between correctly pointed classes and all pointing trials.
As we aim to study class sensitivity we use a subset of PASCAL VOC07~\cite{Everingham15} dataset where there are \emph{multiple object classes present in each image} (as defined in~\cite{zhang2018top}).
\\
\textbf{Restricted pointing game.}
As we are using a dataset with multiple object classes in each image, class-sensitive methods can generally achieve better accuracy in pointing game. However, the major problem with the pointing game experiment is that it does not disentangle class-sensitivity from localization. Therefore, comparing different methods with each other is difficult. 
We propose the restricted pointing game to tackle this problem. 
For each image and all its object classes, we only generate the saliency map for the max output class. The ground truth masks of all objects in the image are all compared against the max saliency map. 
If the resulting accuracy for the restricted pointing game is similar to the accuracy of the original pointing game, it can show that the saliency maps for different object classes are similar and are pointing to the same object class.
\\
\textbf{Class-sensitivity.}
~\cite{Nie2018} showed that for Guided Backpropagation and deconvolution methods, the generated saliency maps for different output classes in the same image are visually indiscernible. 
Motivated by this experiment,~\cite{Rebuffi2020} proposed a quantitative evaluation for class-sensitivity. 
For a given attribution method, saliency maps are generated for the max and min classes, and the correlation between them is computed. We report the average on all images in PASCAL VOC07.
\\
\textbf{Parameter randomization.}
~\cite{Adebayo2018} proposed a sanity check experiment for evaluating the methods’ sensitivity to model parameter randomization. 
The similarity between saliency maps is measured by Spearman on HOG. 
The model parameters are increasingly randomized by reinitializing ($\sim \mathcal{N}(0, 0.01)$) the parameters of layers in a cascading fashion from the last layer to the first. We use a set of 1000 images of ImageNet~\cite{deng2009imagenet}. 

%% file: content/positive_sum.tex
\begin{figure}
    \centering
    \includegraphics[width=\columnwidth]{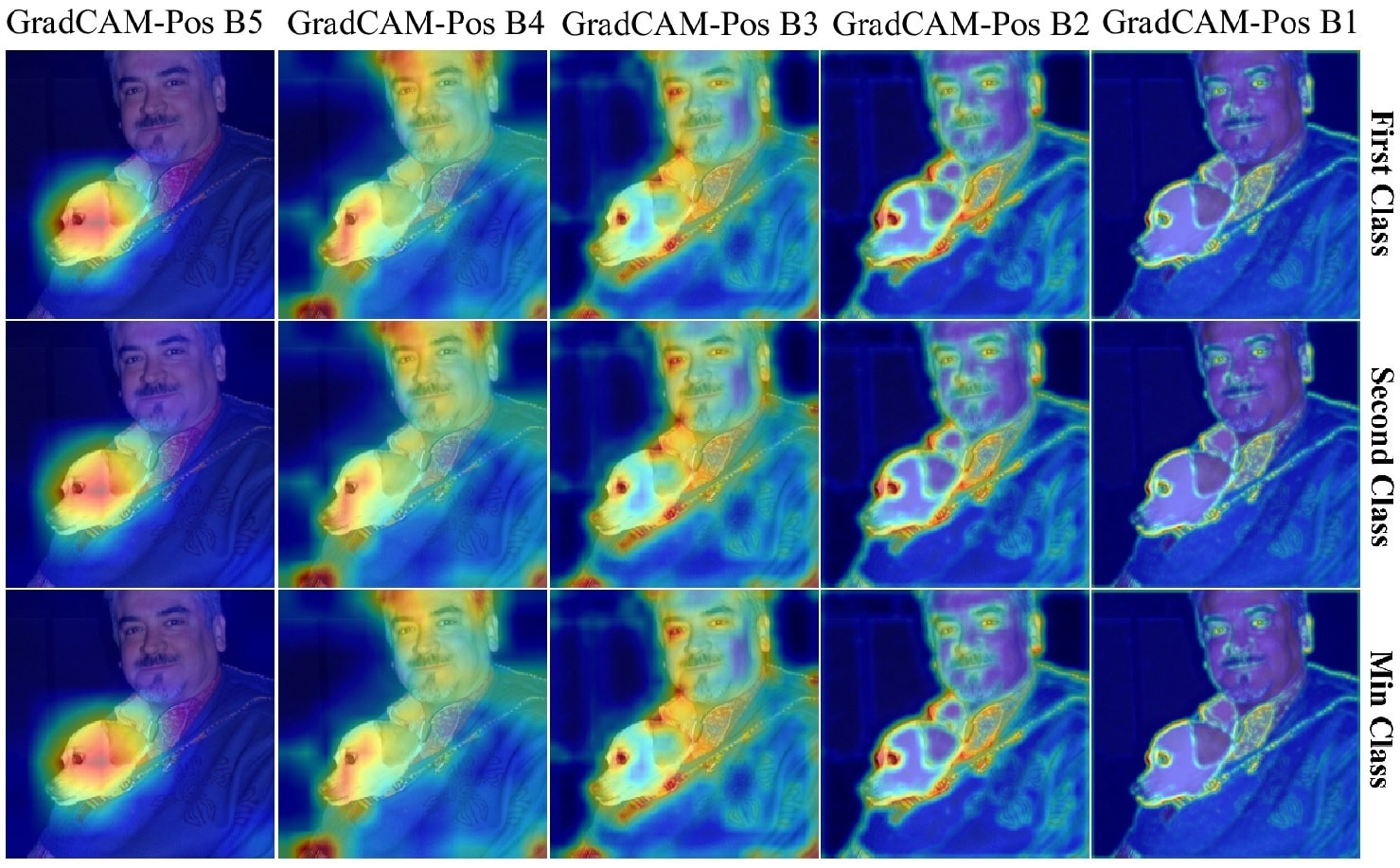}
    \caption{Saliency maps for different output predictions (rows) using GradCAM with positive aggregation (GradCAM\_Pos) on different layers of VGG-16. Left column being the final layer.}
    \label{fig:gradcam_pos}
\end{figure}
The locally connected structure of convolutions results in internal feature maps that retain the spatial information of the input, and therefore an attribution of contributions to these feature maps can be scaled to input space. 
Several methods utilize the gradient of the output with respect to activation units (GradCAM, GradCAM++) or biases (FullGrad) in a chosen layer. 
The gradient information in the chosen layer is aggregated across the channel dimension.
In this section, we show how \textit{positive aggregation} --- using an absolute function (FullGrad) or using $\relu$ (GradCAM++) before the aggregation --- can result in feature maps that recover salient objects in the image without being related to the output prediction being explained.

We also show that simply adding an absolute function to the formulation of GradCAM~\cite{Selvaraju2020} before aggregation can result in the same behavior. 
The positive aggregation allows for GradCAM to be applicable to earlier layers as well, but we show that it is only recovering salient feature information without any regard for the prediction. The original formulation of GradCAM is as follows:
\begin{equation}
S_\text{GradCAM}=\relu(\sum_{k}^{}w_{k}^{}A^{k})
\end{equation}
where $A^{k}$ is the activation map at channel $k$, $f$ is the output being explaind, and
\begin{equation}
w_{k}^{} = \frac{1}{Z}\sum_{i}^{}\sum_{j}^{}\frac{\partial f}{\partial A_{ij}^{k}}
\end{equation}
where $Z$ is the number of activation units in $A^{k}$. 
%
\begin{figure}[t]
    \centering
    \includegraphics[width=\columnwidth]{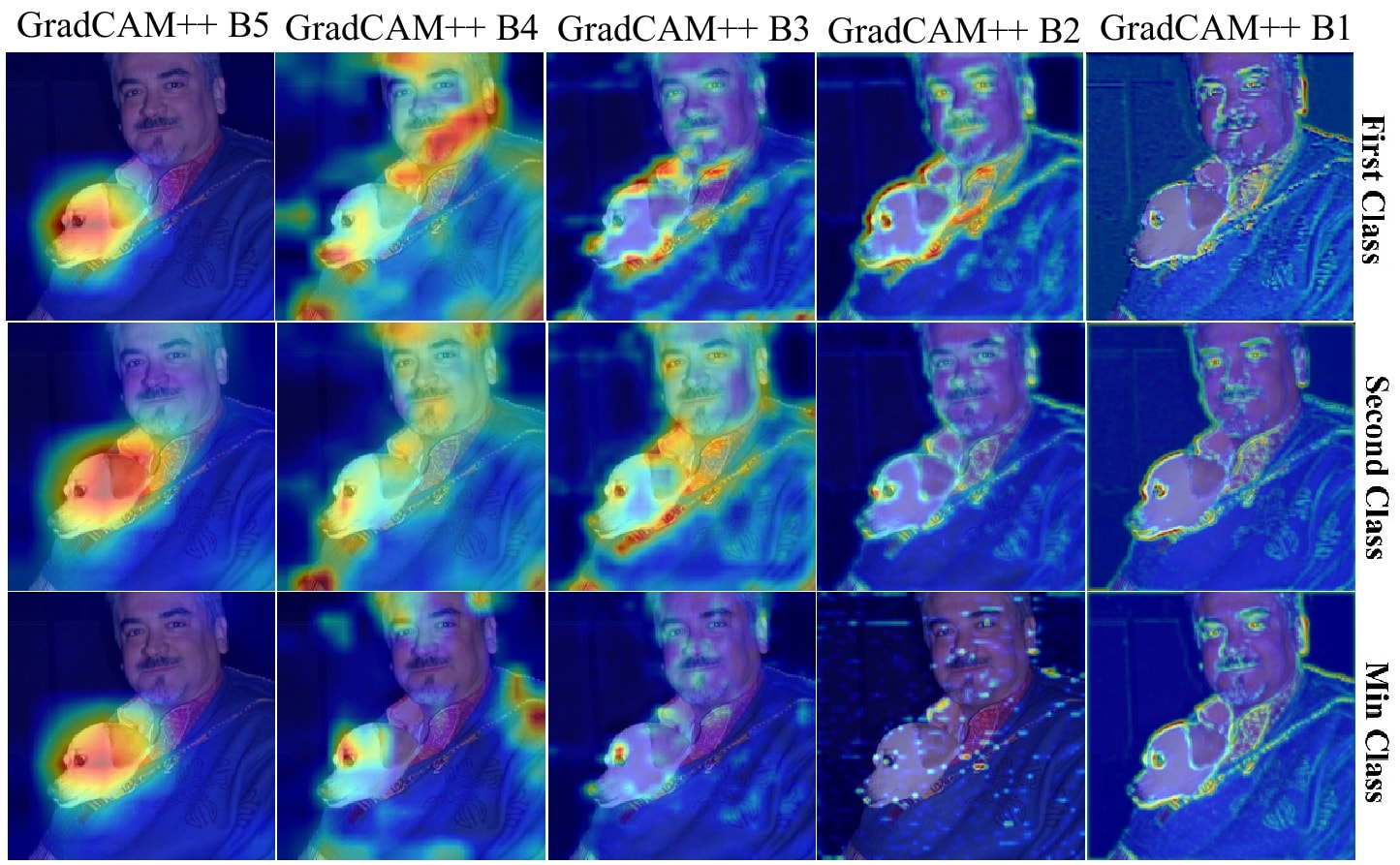}
    \caption{Saliency maps generated for different output predictions (rows) using GradCAM++ on different layers (columns) of VGG-16. Earlier layers presented on the right.}
    \label{fig:gardcam++}
\end{figure}
%
This method is well justified for the final convolutional layer, as such linear approximation of output with respect to activations, which is the basis for the aggregation of feature maps in GradCAM holds stronger for the last layer than previous layers (in a network where CAM~\cite{Zhou2016} is applicable, linear aggregation exactly derives the output).
However, if GradCAM is computed for earlier layers, the resulting saliency maps are arbitrary as reported in~\cite{Selvaraju2020,Rebuffi2020}. 
During the aggregation phase, GradCAM++ uses $\relu$ on gradients before the summation directly in the main formulation of the method as follows:
\begin{equation}
    S_\text{GradCAM++}=\sum_{k}^{}w_{k}^{}A^{k}
\end{equation}
where,
\begin{equation}
w_{k}^{} = \sum_{i}^{}\sum_{j}^{}\alpha _{ij}^{k} \relu(\frac{\partial f}{\partial A_{ij}^{k}})
\end{equation}
and $\alpha _{ij}^{k}$ is a coefficient computed for each activation unit (please refer to~\cite{Chattopadhay2018}).

Fullgrad’s~\cite{Srinivas2019} main formulation distributes the contribution to the output prediction between all inputs to the network (input and mid-level biases) and there is no $\relu$ or absolute summation. However such formulation cannot be visualized as a saliency map. Thus an aggregation phase is added for CNN visualizations, and an absolute function is used within the post-processing $\psi$: 
\begin{equation}
S_\text{FullGrad}= \psi (\frac{\partial f}{\partial x}\odot x )+\sum_{l\in L}^{}\sum_{k}^{}\psi (\frac{\partial f}{\partial A^{k}}\odot b^{l})
\end{equation}
where $x$ is the input image, and $b^{l}$ stands for the bias parameters of layer $l$. The absolute operation is added to visualize only the magnitude of importance while ignoring the sign. However, ignoring the sign results in aggregating the gradients related to all features in the image (compare this to CAM/GradCAM where the summation is done without $|\cdot|$ because the linear combination of activations in the last layer recovers the output). 
\begin{table}[t]
\caption{Pointing game on aggregation methods}
\label{table:pointing_aggregation}
\vskip 0.15in
\begin{center}
\begin{small}
\begin{sc}
\begin{tabular}{lccr}
\toprule
  & Original & Restricted \\
\toprule
GradCAM     &    74.1    &   48.9\\
FullGrad    &    58.4    &   50.5\\
\hline
GradCAM++    &    66.2    &   54.1\\
GradCAM++ B4    &    36.3   &   37.1\\
GradCAM++ B3     &    35.2    &   36.4\\
GradCAM++ B2     &    38.0    &   39.2\\
GradCAM++ B1      &    40.4    &   40.3\\
\hline
GradCAM\_Pos    &    60.4    &   58.1\\
GradCAM\_Pos B4    &    37.6    &   37.8\\
GradCAM\_Pos B3     &    31.8    &   31.6\\
GradCAM\_Pos B2     &    40.6    &   40.7\\
GradCAM\_Pos B1      &    40.7    &   40.6\\
\hline
GradMid    &    63.9    &   56.3\\
GradMid B4    &    65.4    &   56.3\\
GradMid B3     &    60.0    &   53.8\\
GradMid B2     &    53.8    &   48.8\\
GradMid B1      &    51.8    &   44.5\\
\hline
Cumulative\_GradCAM  &    58.1    &   56.2\\
Cumulative\_GradMid  &    64.1    &   56.3\\
\bottomrule
\end{tabular}
\end{sc}
\end{small}
\end{center}
\vskip -0.1in
\end{table}
%

\begin{figure}
    \centering
    \includegraphics[width=\columnwidth]{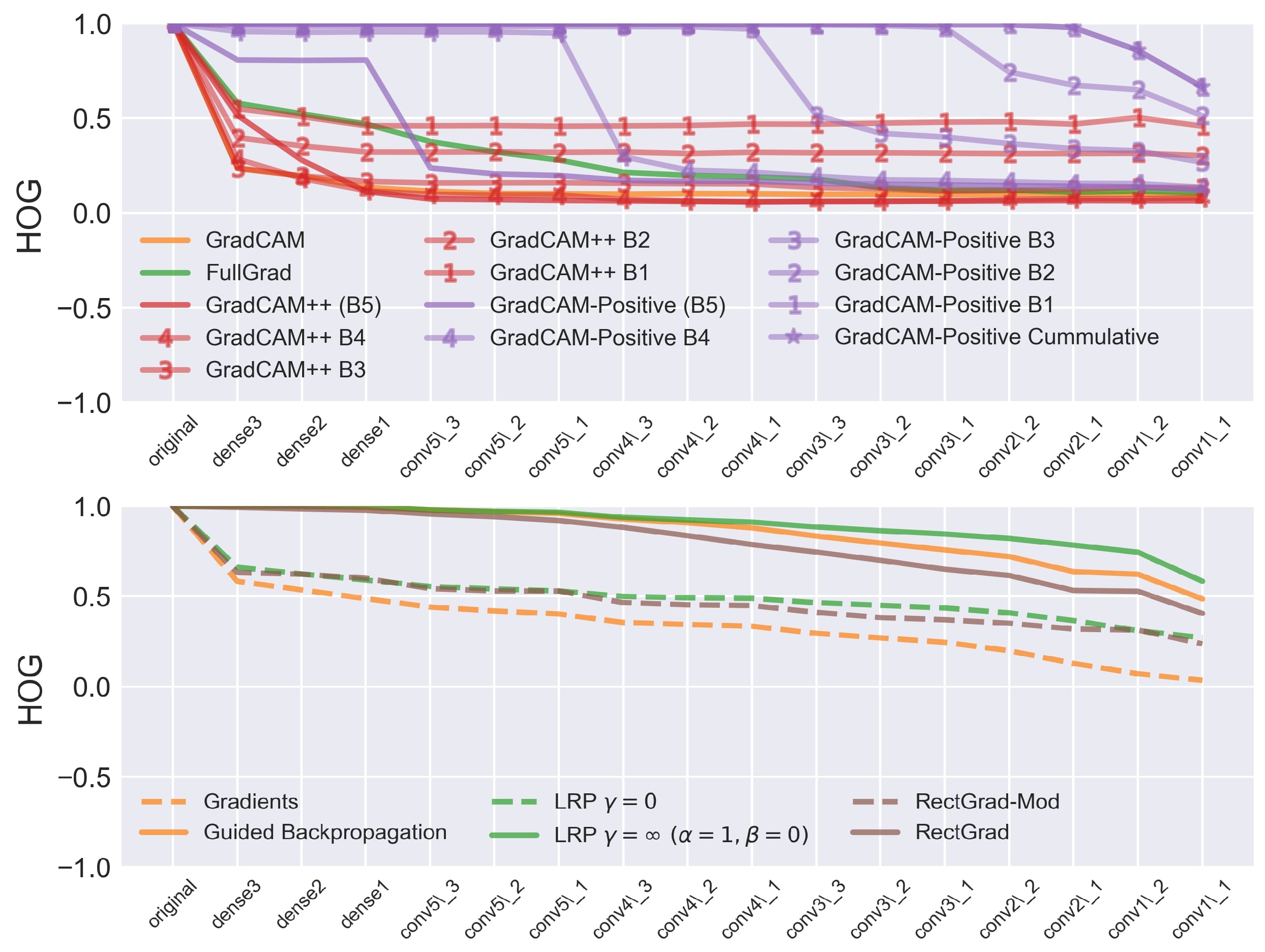}
    \caption{Sanity checks for sensitivity to parameter randomization for aggregation (Up) and propagation (Down) methods. 
    }
    \label{fig:sanity_chart}
\end{figure}

\begin{figure}
    \centering
    \includegraphics[width=\columnwidth]{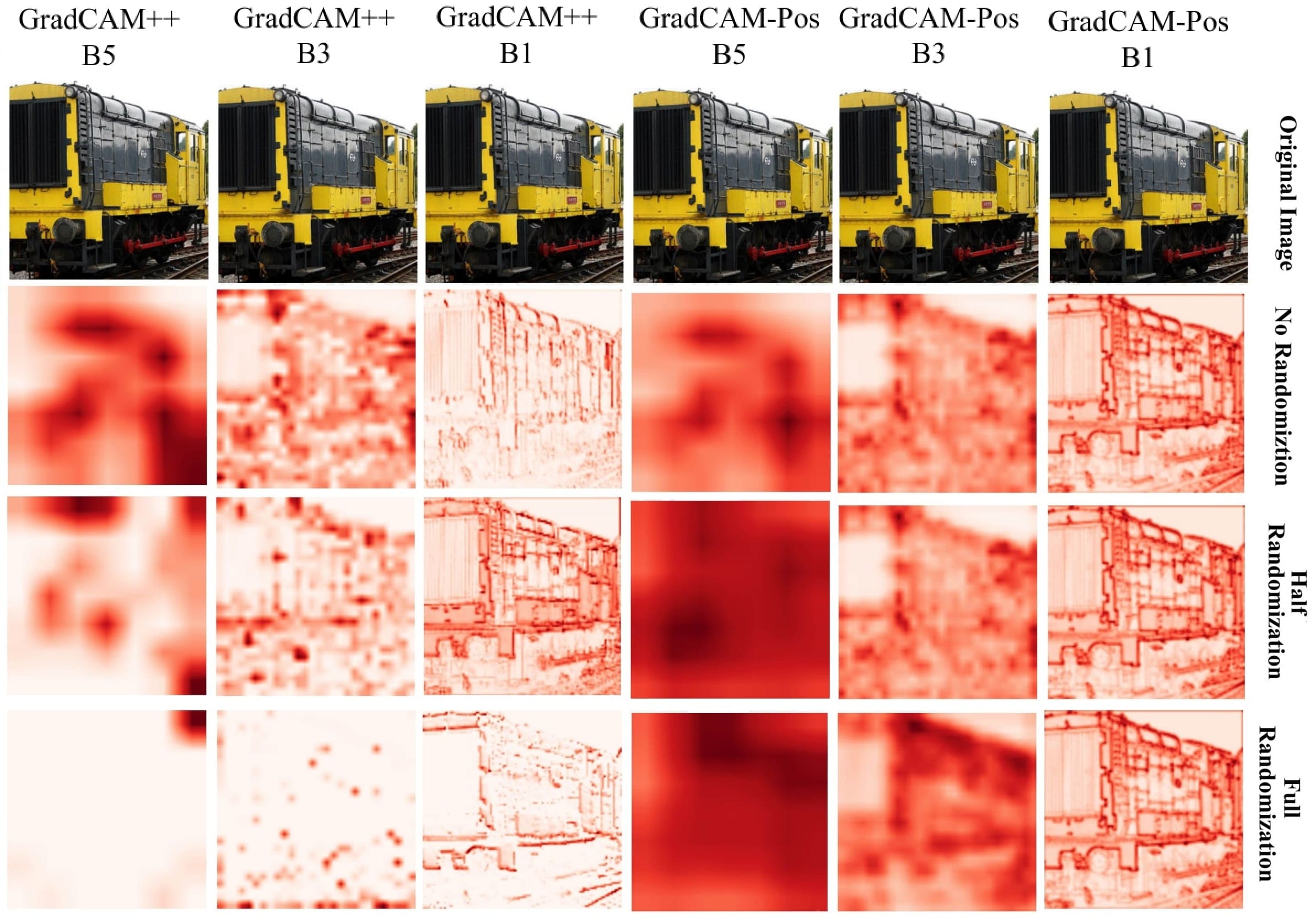}
    \caption{From left: Saliency maps for GradCAM++ B5, B3 and B1 layers. followed by GradCAM\_Pos B5, B3, B1. From top: Original image, saliency map on original model, half randomized model and fully randomized model.}
    \label{fig:sanity_vis}
\end{figure}

Initially, we show that if we change the formulation of GradCAM by adding an absolute function during aggregation, which we call GradCAM\_Pos, the resulting saliency maps for all layers show salient image features, however, it seems that these are merely silhouettes of salient image features and we observe that these features are not relevant to the predicted class. A visual example is presented in Fig.~\ref{fig:gradcam_pos}, ~\ref{fig:aggregation_all_pointing}. 
The maps are computed for different layers of VGG. We use the convolutional maps at different resolution blocks, and represent them as GradCAM\_Pos plus block number, where B1 for instance represents the first block.
It is observed that as we move toward earlier layers the generated maps recover low-level features of objects such as edges. \emph{The maps on earlier layers should not be interpreted as 'high resolution', as they are not attribution or saliency maps anymore.} It is also reported by~\cite{Rebuffi2020} that earlier layers result in class-insensitive maps in aggregation methods, however, we are considering the role positive aggregation plays. \emph{With positive aggregation, even in the final layer, maps highlight salient image features but for both classes present in the image.} This observation is further supported in the pointing game experiments in Table~\ref{table:pointing_aggregation} as the original and restricted games accuracies are similar. 
Class-sensitivity metric in Table~\ref{table:corr_aggregation} shows that for all layers, except the final layer, the correlation is close to 1. For the final layer the correlation is lower but still \emph{significantly higher than GradCAM with absolute function (0.69 vs. 0.03), showing how detrimental the effect of positive aggregation is.} The positive aggregation in this case also has a significant effect on the method's sensitivity to randomization. Fig.~\ref{fig:sanity_vis} and ~\ref{fig:sanity_chart} shows that \emph{these method's saliency maps do not change after randomization, further supporting the claim that methods are recovering image features.}

As stated, the ability to generate saliency maps that look interpretable on earlier layers for GradCAM can be achieved via positive aggregation. GradCAM++ also uses positive aggregation and its visual examples for the last layer and preceding layers are presented in Fig \ref{fig:gardcam++} and Fig.~\ref{fig:aggregation_all_pointing}. The observations for GradCAM\_Pos are visible here as well. The results on the restricted pointing game and original are equivalent when experiments are done on all layers except the final layer. However, correlations (Fig. ~\ref{table:corr_aggregation}) are better than the GradCAM\_Pos case, but still significantly high compared to GradCAM. These lower correlations are expected when looking at the visual results as there are arbitrary highlighted areas in the maps as well, but the method still highlights all the salient features in the image, which is confirmed by pointing game experiments (Table ~\ref{table:pointing_aggregation}). In parameter randomization in Fig.~\ref{fig:sanity_chart} and ~\ref{fig:sanity_vis} \emph{the method gets more insensitive when applied to earlier layers.}
\begin{table}
\caption{Class-sensitivity metric on aggregation methods}
\label{table:corr_aggregation}
\vskip 0.15in
\begin{center}
\begin{small}
\begin{sc}
\begin{tabular}{lccr}
\toprule
  & Spearman HOG\\
\toprule
GradCAM     &    0.03    \\
FullGrad    &    0.48    \\
\hline
GradCAM++    &    0.37    \\
GradCAM++ B4    &    0.23  \\
GradCAM++ B3     &    0.44    \\
GradCAM++ B2     &    0.58    \\
GradCAM++ B1      &    0.65    \\
\hline
GradCAM\_Pos    &    0.69    \\
GradCAM\_Pos B4    &    0.94    \\
GradCAM\_Pos B3     &   0.98    \\
GradCAM\_Pos B2     &   0.99    \\
GradCAM\_Pos B1      &  0.99    \\
\hline
GradMid    &    0.29    \\
GradMid B4    &    0.35    \\
GradMid B3     &   0.32    \\
GradMid B2     &   0.34    \\
GradMid B1      &  0.43    \\
\hline
Cumulative\_GradCAM  &    0.92    \\
CumulativeMid  &    0.36    \\
\bottomrule
\end{tabular}
\end{sc}
\end{small}
\end{center}
\vskip -0.1in
\end{table}

Fullgrad has one major difference in formulation compared to the other methods. The method has an extra aggregation step where maps from all \emph{layers} are aggregated. It is expected that is aggregations compounds the destructive effect of positive aggregation within each layer. In this section, we study the effect of aggregation on all layers. First, we propose an all-layer aggregation on GradCAM\_Pos and call the method CumulativeGradCAM. As can be seen in Table ~\ref{table:corr_aggregation} and ~\ref{table:pointing_aggregation} the method is not class sensitive. The result seems to be bounded by the worst-case early layer and best case final layer for GradCAM\_Pos. This also explains why FullGrad is less class sensitive than GradCAM++.
CumulativeGradCAM uses the model's activations, therefore we also propose a variant where we only aggregate gradients. When done on one layer we call this as GradMid and when done on all layers we call it CumuluativeGradMid. This is to show that \emph{positive aggregation on gradients alone without any weighting can also generate visually interpretable feature maps, though similar to other cases the visualization only shows salient image features irrespective of the model and the output.} The results for these all-layer summation methods are presented alongside all the aforementioned methods in Fig.~\ref{fig:aggregation_all_pointing}. \emph{It is also observed in Fig.~\ref{fig:sanity_chart} that Fullgrad is insensitive to randomization considerably, and the CumulativeGradCAM is the least sensitive.}
%
%
%
%
%
%
%
%

%% file: content/positive_prop.tex
\begin{table}[t]
\caption{Pointing game on propagation methods}
\label{table:pointing_prop}
\vskip 0.15in
\begin{center}
\begin{small}
\begin{sc}
\begin{tabular}{lccr}
\toprule
  & Original & Restricted \\
\toprule
Guided BackProp    &    48.7    &   47.5\\
Gradients     &    52.7    &   45.1\\
\hline
RectGrad    &    51.2    &   50.0\\
RectGrad\_mod    &    59.5   &   53.8\\
\hline
LRP-$\gamma=\infty$     &    50.6    &   50.0\\
LRP-0     &    51.1    &   44.2\\
\bottomrule
\end{tabular}
\end{sc}
\end{small}
\end{center}
\vskip -0.1in
\end{table}

\begin{figure*}
    \centering
    \includegraphics[width=0.8\textwidth]{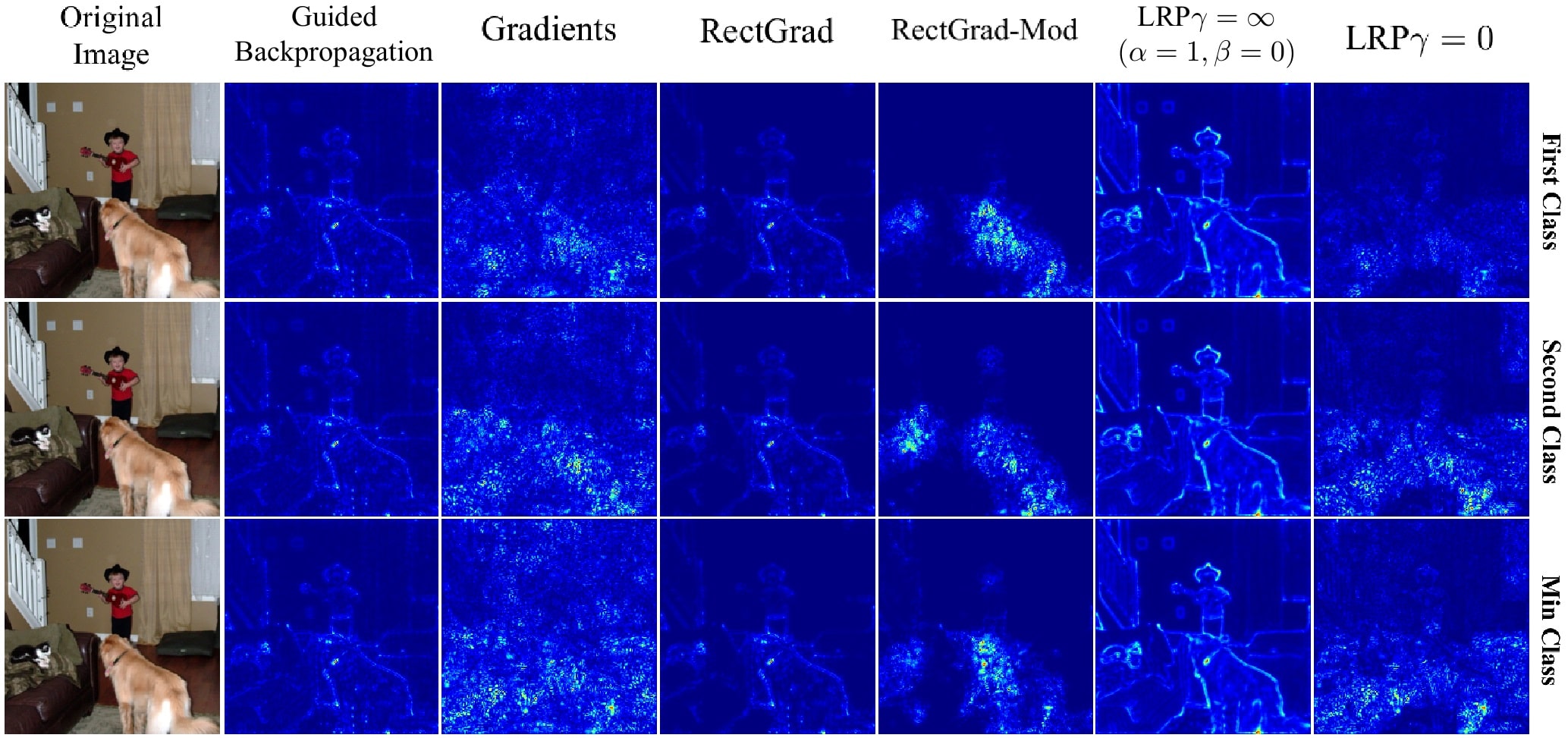}
    \caption{Saliency maps generated for different output predictions (rows) using different propagation methods.}
    \label{fig:propgation_all}
\end{figure*}

Gradient information backpropagation towards the input from the the output being explained is the underlying principle of a class of gradient-based saliency methods.
In this section we study the effect of rules that propagate only positive gradient information. Considering a ReLU based neural network,
let $a_{i}^{l}$ denote an activation unit $i$ in layer $l$ and the flowing gradient into $a_{i}^{l}$ be $R_{i}^{l+1}$ and let the gradient backpropagated from $a_{i}^{l}$ be $R_{i}^{l}$. In normal gradient backpropagation $R_{i}^{l} = \mathbb{I}(a_{i}^{l})R_{i}^{l+1}$, where $\mathbb{I}(.)$ is the indicator function. Guided Backpropagation rule is defined as:
\begin{equation}
R_{i}^{l} = \mathbb{I}(a_{i}^{l}R_{i}^{l+1}>0)R_{i}^{l+1}
\end{equation}
and RectGrad uses the following propagation rule:
\begin{equation}
R_{i}^{l} = \mathbb{I}(a_{i}^{l}R_{i}^{l+1}>\tau )R_{i}^{l+1}
\end{equation}
where $\tau$ is a threshold value. 


Layerwise Relevance Propagation (LRP) has several propagation rules and a widely adopted rule is the LRP-$\alpha1\beta0$, which is equivalent~\cite{Samek2019} to Deep Taylor $\alpha1\beta0$~\cite{Montavon2017}, LRP z+- rule and Excitation-Backprop~\cite{zhang2018top} methods. This rule is also equivalent to LRP-$\gamma$ with $\gamma=\infty$. LRP-$\gamma$ is introduced to favor the effect of positive contributions over negative ones and is defined as follows:

\begin{equation}
    R_{j} = \sum_{k}^{}\frac{a_{j}(w_{jk}+\gamma w_{jk}^{+})}{\sum_{j}^{}a_{j}(w_{jk}+\gamma w_{jk}^{+})}R_{k}
\end{equation}

where $R_{j}$ is the relevance of neuron $j$ for prediction and $R_{k}$ is relevance of neuron $k$ in the next layer. The controlling parameter for considering positive contributions is $\gamma$. As $\gamma$ is increased the effect is more pronounced. In this work we consider the limits $\gamma=\infty$ and $\gamma=0$ (LRP-0).
\begin{table}[t]
\caption{Class-sensitivity metric on propagation methods}
\label{table:corr_prop}
\vskip 0.15in
\begin{center}
\begin{small}
\begin{sc}
\begin{tabular}{lccr}
\toprule
  & Spearman HOG \\
\toprule
Guided BackProp    &    0.99    \\
Gradients     &    0.44    \\
\hline
RectGrad    &    0.99    \\
RectGrad\_Mod    &    0.72   \\
\hline
LRP-$\gamma=\infty$     &    0.99    \\
LRP-0     &    0.59    \\
\bottomrule
\end{tabular}
\end{sc}
\end{small}
\end{center}
\vskip -0.1in
\end{table}
For all these methods, we show the effect of positive gradient information backpropagation by running the experiments in section~\ref{sec:exp} on these methods and their counterpart versions where there is no bias towards positive gradients. The counterpart version for Guided Backpropagation is changing the rule to $R_{i}^{l} = \mathbb{I}(a_{i}^{l}>0)R_{i}^{l+1}$ so that both positive and negative information is propagated and evidently this is equivalent to unmodified gradients. The counterpart version for RectGrad is achievable by using an absolute function in the formulation so that large positive and negative gradients can both flow backwards, therefore the rule is modified to $R_{i}^{l} = \mathbb{I}(\left | a_{i}^{l}R_{i}^{l+1} \right |>\tau )R_{i}^{l+1}$, which we call RectGrad\_Mod. Setting $\gamma=0$ in LRP-$\gamma$ removes the bias towards positive information backpropagation.

We first investigate the effect of positive backpropagation on the sensitivity of the methods to model parameter randomization. ~\cite{Adebayo2018,Nie2018} show that the Guided Backpropagation method is insensitive to parameter randomization. \cite{khakzar2019explaining} report insensitivity to randomization of RectGrad using sanity checks. Our sanity check experiment reported in Fig.~\ref{fig:sanity_chart} further shows that \emph{RectGrad and LRP-$\gamma=\infty$ are also insensitive to the same extent as Guided Backpropagation}. As the resulting saliency maps resemble the images before and after randomization, it points to the fact that such \emph{positive gradient information backpropagation recovers salient features in the image regardless of the model}. The results for the counterpart versions of all methods signify that these counterpart methods are sensitive to randomization.

The pointing game experiment in Table~\ref{table:pointing_prop} shows that the resulting accuracy on the original is similar to accuracy on restricted pointing games are for the positive propagation methods. \emph{This signifies that these methods are pointing to the same salient object class in the images for different output predictions}. On the other hand, for the counterpart versions, the accuracy drops in the restricted version, implying that indeed the saliency maps for different outputs differ. However, it still seems that these counterpart methods (e.g. Gradients) are also class-insensitive for many images (Fig.~\ref{fig:propgation_all}), and effect which needs to be studied in future, especially when other works \cite{Rebuffi2020,Nie2018} have contradicting statements in this regard.

The class-sensitivity metric in Table~\ref{table:corr_prop} shows that there is a significant similarity between saliency maps of different outputs for the positive backpropagation methods. This further confirms the observation that \emph{these methods are recovering salient image features rather than explaining the output prediction.} It is also observed that for counterpart methods, the saliency maps change for different output predictions (Table \ref{table:pointing_prop}, ~\ref{table:corr_prop}), though it does not mean that they are class sensitive (Fig.~\ref{fig:propgation_all}). \emph{Nevertheless, positive propagation makes these methods class-insensitive.}




%% file: content/conclusion.tex
In this work, we empirically showed that positive aggregation or propagation of gradients in gradient-based saliency methods results in saliency maps that recover salient image features regardless of the model and output prediction, and changes the methods towards being class-insensitive and insensitive to randomization.

%% file: main.bbl
\begin{thebibliography}{23}
\providecommand{\natexlab}[1]{#1}
\providecommand{\url}[1]{\texttt{#1}}
\expandafter\ifx\csname urlstyle\endcsname\relax
  \providecommand{\doi}[1]{doi: #1}\else
  \providecommand{\doi}{doi: \begingroup \urlstyle{rm}\Url}\fi

\bibitem[Adebayo et~al.(2018)Adebayo, Gilmer, Muelly, Goodfellow, Hardt, and
  Kim]{Adebayo2018}
Adebayo, J., Gilmer, J., Muelly, M., Goodfellow, I., Hardt, M., and Kim, B.
\newblock {Sanity checks for saliency maps}.
\newblock In \emph{Advances in Neural Information Processing Systems}, 2018.

\bibitem[Bach et~al.(2015)Bach, Binder, Montavon, Klauschen, M{\"{u}}ller, and
  Samek]{Bach2015}
Bach, S., Binder, A., Montavon, G., Klauschen, F., M{\"{u}}ller, K.~R., and
  Samek, W.
\newblock {On pixel-wise explanations for non-linear classifier decisions by
  layer-wise relevance propagation}.
\newblock \emph{PLoS ONE}, 2015.
\newblock ISSN 19326203.
\newblock \doi{10.1371/journal.pone.0130140}.

\bibitem[Chattopadhay et~al.(2018)Chattopadhay, Sarkar, Howlader, and
  Balasubramanian]{Chattopadhay2018}
Chattopadhay, A., Sarkar, A., Howlader, P., and Balasubramanian, V.~N.
\newblock {Grad-CAM++: Generalized gradient-based visual explanations for deep
  convolutional networks}.
\newblock In \emph{Proceedings - 2018 IEEE Winter Conference on Applications of
  Computer Vision, WACV 2018}, 2018.
\newblock ISBN 9781538648865.
\newblock \doi{10.1109/WACV.2018.00097}.

\bibitem[Deng et~al.(2009)Deng, Dong, Socher, Li, Li, and
  Fei-Fei]{deng2009imagenet}
Deng, J., Dong, W., Socher, R., Li, L.-J., Li, K., and Fei-Fei, L.
\newblock Imagenet: A large-scale hierarchical image database.
\newblock In \emph{2009 IEEE conference on computer vision and pattern
  recognition}, pp.\  248--255. Ieee, 2009.

\bibitem[Everingham et~al.(2015)Everingham, Eslami, Van~Gool, Williams, Winn,
  and Zisserman]{Everingham15}
Everingham, M., Eslami, S. M.~A., Van~Gool, L., Williams, C. K.~I., Winn, J.,
  and Zisserman, A.
\newblock The pascal visual object classes challenge: A retrospective.
\newblock \emph{International Journal of Computer Vision}, 111\penalty0
  (1):\penalty0 98--136, January 2015.

\bibitem[Fong et~al.(2019)Fong, Patrick, and Vedaldi]{fong2019understanding}
Fong, R., Patrick, M., and Vedaldi, A.
\newblock Understanding deep networks via extremal perturbations and smooth
  masks.
\newblock In \emph{Proceedings of the IEEE International Conference on Computer
  Vision}, pp.\  2950--2958, 2019.

\bibitem[Hooker et~al.(2019)Hooker, Erhan, Kindermans, and
  Kim]{hooker2019benchmark}
Hooker, S., Erhan, D., Kindermans, P.-J., and Kim, B.
\newblock A benchmark for interpretability methods in deep neural networks.
\newblock In \emph{Advances in Neural Information Processing Systems}, pp.\
  9737--9748, 2019.

\bibitem[Khakzar et~al.(2019)Khakzar, Baselizadeh, Khanduja, Kim, and
  Navab]{khakzar2019explaining}
Khakzar, A., Baselizadeh, S., Khanduja, S., Kim, S.~T., and Navab, N.
\newblock Explaining neural networks via perturbing important learned features.
\newblock \emph{arXiv preprint arXiv:1911.11081}, 2019.

\bibitem[Kim et~al.(2020)Kim, Seo, Jeon, Koo, Choe, and Jeon]{Kim2020}
Kim, B., Seo, J., Jeon, S., Koo, J., Choe, J., and Jeon, T.
\newblock {Why are Saliency Maps Noisy? Cause of and Solution to Noisy Saliency
  Maps}.
\newblock 2020.
\newblock \doi{10.1109/iccvw.2019.00510}.

\bibitem[Kindermans et~al.(2019)Kindermans, Hooker, Adebayo, Alber,
  Sch{\"{u}}tt, D{\"{a}}hne, Erhan, and Kim]{Kindermans2019}
Kindermans, P.~J., Hooker, S., Adebayo, J., Alber, M., Sch{\"{u}}tt, K.~T.,
  D{\"{a}}hne, S., Erhan, D., and Kim, B.
\newblock {The (Un)reliability of Saliency Methods}.
\newblock In \emph{Lecture Notes in Computer Science (including subseries
  Lecture Notes in Artificial Intelligence and Lecture Notes in
  Bioinformatics)}. 2019.
\newblock \doi{10.1007/978-3-030-28954-6_14}.

\bibitem[Lundberg \& Lee(2017)Lundberg and Lee]{Lundberg2017}
Lundberg, S.~M. and Lee, S.~I.
\newblock {A unified approach to interpreting model predictions}.
\newblock In \emph{Advances in Neural Information Processing Systems}, 2017.

\bibitem[Montavon et~al.(2017)Montavon, Lapuschkin, Binder, Samek, and
  M{\"{u}}ller]{Montavon2017}
Montavon, G., Lapuschkin, S., Binder, A., Samek, W., and M{\"{u}}ller, K.~R.
\newblock {Explaining nonlinear classification decisions with deep Taylor
  decomposition}.
\newblock \emph{Pattern Recognition}, 2017.
\newblock ISSN 00313203.
\newblock \doi{10.1016/j.patcog.2016.11.008}.

\bibitem[Nie et~al.(2018)Nie, Zhang, and Patel]{Nie2018}
Nie, W., Zhang, Y., and Patel, A.~B.
\newblock {A theoretical explanation for perplexing behaviors of
  backpropagation-based visualizations}.
\newblock In \emph{35th International Conference on Machine Learning, ICML
  2018}, 2018.
\newblock ISBN 9781510867963.

\bibitem[Rebuffi et~al.(2020)Rebuffi, Fong, Ji, and Vedaldi]{Rebuffi2020}
Rebuffi, S.-A., Fong, R., Ji, X., and Vedaldi, A.
\newblock {There and Back Again: Revisiting Backpropagation Saliency Methods}.
\newblock apr 2020.
\newblock URL \url{http://arxiv.org/abs/2004.02866}.

\bibitem[Samek et~al.(2019)Samek, Montavon, Vedaldi, Hansen, and
  Muller]{Samek2019}
Samek, W., Montavon, G., Vedaldi, A., Hansen, L.~K., and Muller, K.-R.
\newblock {Explainable AI: Interpreting, Explaining and Visualizing Deep
  Learning}.
\newblock \emph{Lecture Notes in Computer Science}, 2019.
\newblock \doi{10.1007/978-3-030-28954-6}.

\bibitem[Selvaraju et~al.(2020)Selvaraju, Cogswell, Das, Vedantam, Parikh, and
  Batra]{Selvaraju2020}
Selvaraju, R.~R., Cogswell, M., Das, A., Vedantam, R., Parikh, D., and Batra,
  D.
\newblock {Grad-CAM: Visual Explanations from Deep Networks via Gradient-Based
  Localization}.
\newblock \emph{International Journal of Computer Vision}, 2020.
\newblock ISSN 15731405.
\newblock \doi{10.1007/s11263-019-01228-7}.

\bibitem[Simonyan \& Zisserman(2014)Simonyan and Zisserman]{simonyan2014very}
Simonyan, K. and Zisserman, A.
\newblock Very deep convolutional networks for large-scale image recognition.
\newblock \emph{arXiv preprint arXiv:1409.1556}, 2014.

\bibitem[Springenberg et~al.(2015)Springenberg, Dosovitskiy, Brox, and
  Riedmiller]{Springenberg2015}
Springenberg, J.~T., Dosovitskiy, A., Brox, T., and Riedmiller, M.
\newblock {Striving for simplicity: The all convolutional net}.
\newblock In \emph{3rd International Conference on Learning Representations,
  ICLR 2015 - Workshop Track Proceedings}, 2015.

\bibitem[Srinivas \& Fleuret(2019)Srinivas and Fleuret]{Srinivas2019}
Srinivas, S. and Fleuret, F.
\newblock {Full-Gradient Representation for Neural Network Visualization}.
\newblock Technical report, 2019.

\bibitem[Sundararajan \& Najmi(2019)Sundararajan and
  Najmi]{sundararajan2019many}
Sundararajan, M. and Najmi, A.
\newblock The many shapley values for model explanation.
\newblock \emph{arXiv preprint arXiv:1908.08474}, 2019.

\bibitem[Sundararajan et~al.(2017)Sundararajan, Taly, and
  Yan]{Sundararajan2017}
Sundararajan, M., Taly, A., and Yan, Q.
\newblock {Axiomatic attribution for deep networks}.
\newblock In \emph{34th International Conference on Machine Learning, ICML
  2017}, 2017.
\newblock ISBN 9781510855144.

\bibitem[Zhang et~al.(2018)Zhang, Bargal, Lin, Brandt, Shen, and
  Sclaroff]{zhang2018top}
Zhang, J., Bargal, S.~A., Lin, Z., Brandt, J., Shen, X., and Sclaroff, S.
\newblock Top-down neural attention by excitation backprop.
\newblock \emph{International Journal of Computer Vision}, 126\penalty0
  (10):\penalty0 1084--1102, 2018.

\bibitem[Zhou et~al.(2016)Zhou, Khosla, Lapedriza, Oliva, and
  Torralba]{Zhou2016}
Zhou, B., Khosla, A., Lapedriza, A., Oliva, A., and Torralba, A.
\newblock {Learning Deep Features for Discriminative Localization}.
\newblock In \emph{Proceedings of the IEEE Computer Society Conference on
  Computer Vision and Pattern Recognition}, 2016.
\newblock ISBN 9781467388504.
\newblock \doi{10.1109/CVPR.2016.319}.

\end{thebibliography}
